\documentclass[letterpaper]{article} 
\usepackage{aaai2026}  
\usepackage{times}  
\usepackage{helvet}  
\usepackage{courier}  
\usepackage[hyphens]{url}  
\usepackage{graphicx} 
\usepackage{natbib}  
\usepackage{caption} 
\usepackage{algorithm}
\usepackage{algorithmic}
\usepackage{svg}
\usepackage{colortbl,hhline}

\usepackage{newfloat}
\usepackage{listings}
\DeclareCaptionStyle{ruled}{labelfont=normalfont,labelsep=colon,strut=off} 
\floatstyle{ruled}
\newfloat{listing}{tb}{lst}{}
\floatname{listing}{Listing}
\title{Deploying Rapid Damage Assessments from sUAS Imagery for Disaster Response}
\author{
    Thomas Manzini, Priyankari Perali, Robin R. Murphy
}
\affiliations{
    Texas A\&M University, Department of Computer Science and Engineering\\
    L.F. Peterson Building, 435 Nagle St, College Station, TX 77843\\
    tmanzini@tamu.edu, perali@tamu.edu, robin.r.murphy@tamu.edu
}

\begin{document}

\maketitle

\begin{abstract}
This paper presents the first AI/ML system for automating building damage assessment in uncrewed aerial systems (sUAS) imagery to be deployed operationally during federally declared disasters (Hurricanes Debby and Helene). In response to major disasters, sUAS teams are dispatched to collect imagery of the affected areas to assess damage; however, at recent disasters, teams collectively delivered between 47GB and 369GB of imagery per day, representing more imagery than can reasonably be transmitted or interpreted by subject matter experts in the disaster scene, thus delaying response efforts. To alleviate this data avalanche encountered in practice, computer vision and machine learning techniques are necessary. While prior work has been deployed to automatically assess damage in satellite imagery, there is no current state of practice for sUAS-based damage assessment systems, as all known work has been confined to academic settings. 
This work establishes the state of practice via the development and deployment of models for building damage assessment with sUAS imagery. 
The model development involved training on the largest known dataset of post-disaster sUAS aerial imagery, containing 21,716 building damage labels, and the operational training of 91 disaster practitioners. The best performing model was deployed during the responses to Hurricanes Debby and Helene, where it assessed a combined 415 buildings in approximately 18 minutes.
This work contributes documentation of the actual use of AI/ML for damage assessment during a disaster and lessons learned to the benefit of the AI/ML research and user communities.
\end{abstract}

\section{Introduction}

\begin{figure}
    \centering
    \includegraphics[width=0.9\linewidth]{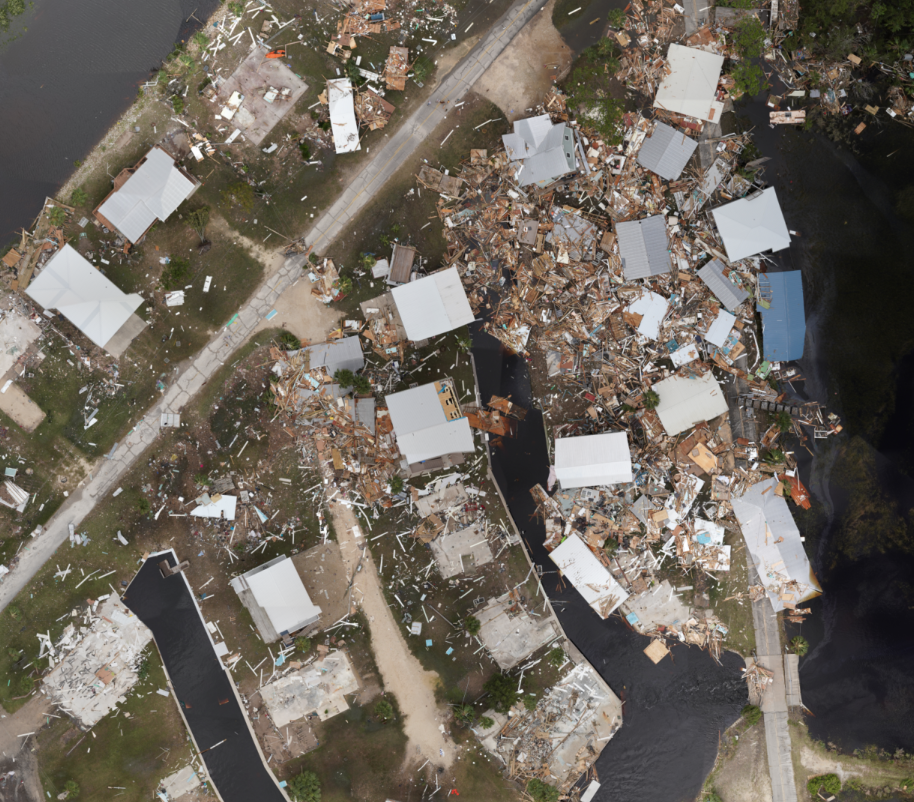}
    \caption{sUAS imagery used by the AI/ML system to automatically assess building damage. Imagery from Dekle Beach, FL, USA, after Hurricane Helene. Credit FL-UAS1.}
    \label{fig:sample_imagery}
\end{figure}

\begin{figure*}
    \centering
    \includegraphics[width=\linewidth]{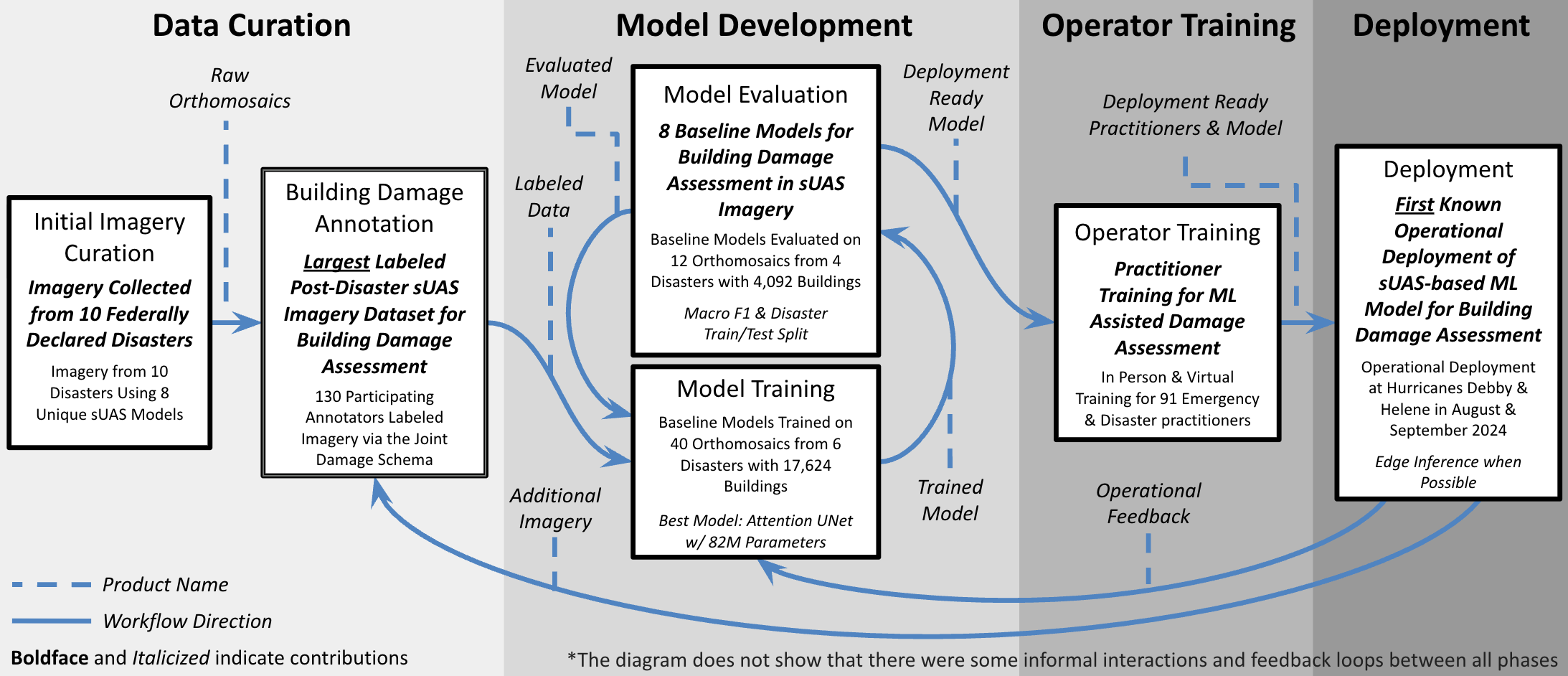}
    \caption{The overall development cycle used in this work, with the Data Curation stage detailed in Sec. ``Data Curation," Model Development, and Operator Training are detailed in Sec. ``Approach" and Deployment is detailed in Sec. ``Initial Deployments at Hurricanes Debby \& Helene."}
    \label{fig:deployment_cycle}
\end{figure*}

When a disaster occurs, disaster response operations deploy multiple sensors to capture imagery of the impacted areas to aid decision-making. 
Recently, small uncrewed aerial systems (sUAS) have become more commonplace \cite{lozano2023data, manzini2023quantitative, fernandes2018quantitative, fernandes2019quantitative}.
The motivation behind the increased use of sUAS to capture imagery is due to their high resolution and relative ease of deployment compared to other sensors, such as crewed aircraft and satellites. A sample of sUAS imagery is shown in Fig. \ref{fig:sample_imagery}. One of the uses for the imagery collected is damage assessments, which inform the emergency managers of where the affected areas are and can improve the allocation and navigation of aid \cite{fema2025fema}. 

One major challenge with sUAS is the overwhelming amount of collected data. 
At recent disasters, sUAS teams delivered between 47GB and 369GB of imagery per day, which often had to be physically transported to decision makers instead of transmitted wirelessly \cite{manzini2023quantitative, manzini2023wireless}. Upon arrival, this creates a data avalanche, resulting in an impractical volume of data for subject matter experts to assess. 

Machine learning (ML) has been proposed as a potential mechanism to mitigate this data avalanche \cite{akter2019big, garcia2021computer, manzini2023wireless, manzini2023harnessing, manzini2023quantitative}. 
By offloading analysis of imagery onto ML algorithms running on compute deployed in the field where the data is collected, and transmitting smaller, imagery-derived ML-based data products, decision-makers could leverage those data products instead \cite{manzini2025challenges}. Theoretically, this would enable decision-making at least one day earlier than the status quo, where decision makers inspect imagery at least one day after collection due to a combination of transmission delays and sleep cycles \cite{manzini2023wireless, manzini2023quantitative}. Such an improvement in decision times is expected to have an out-sized impact on both response and recovery efforts, leading to decreases in individual and economic consequences for the disaster-affected areas \cite{haas1977reconstruction, murphy2010national}.

Despite this potential, there are no known efforts to operationalize or deploy any of the academic work on sUAS imagery, as all known efforts have been confined to academia \cite{rahnemoonfar2023rescuenet, rahnemoonfar2021floodnet, cheng2021dorianet, zhu2021msnet, pi2020convolutional}. 
Previous efforts in the operational deployment of AI/ML systems for aerial imagery in disaster response operations have been limited to satellite imagery \cite{robinson2023turkey, robinson2023rapid}. However, AI/ML systems for damage assessment with satellite imagery cannot be used for sUAS imagery for two reasons. First, sUAS imagery is at a higher resolution (2-5cm/px) versus satellite imagery (30-50cm/px). Second, even if sUAS imagery was downsampled, labels derived from sUAS imagery and satellite imagery represent significantly different distributions, suggesting it would be inappropriate to deploy models trained on one source against another \cite{manzini2025now}. 

This paper presents the first AI/ML system for automating building damage assessment in sUAS imagery to be deployed during federally declared disasters (Hurricanes Debby and Helene), enabling future deployments of AI/ML systems for disaster response operations.  
The lack of documented deployment of sUAS-based AI/ML systems for disaster response operations deprives the AI/ML communities of the appropriate development cycle they should employ, limiting the understanding of the real-world challenges. While this work represents the first operational deployment, meaningful metrics regarding the impacts of the AI/ML system have not yet been captured, indicating that a more comprehensive deployment is still required. 


The development of the AI/ML system consists of four phases: data curation, model development, operator training, and deployment, each of which offers novel contributions to the AI/ML research and user communities. 
In order of value, the deployment is of particular interest as it documents the deployment cycle that is expected to be used as a template for future efforts in deploying AI/ML systems for disaster response operations. 
Model development contributes a comparison of eight baseline models, employing macro F1 and test/train split to evaluate the model's ability to generalize to unseen disasters, and finds that an Attention UNet was the most performant. Findings from the operator training contribute to an understanding of the disaster practitioners' considerations when deploying such systems. Data curation produced the largest labeled post-disaster sUAS imagery dataset for building damage assessment. As a meta-contribution, the four stages taken together form a template suitable for future efforts by the larger community for disaster response applications. Fig. 2 provides an anchoring overview and general framework for the remainder of the paper.

\section{Related Work}

A survey of literature from the AI/ML, robotics, and emergency management literature indicates that this work is the only operational deployment of an sUAS-based AI/ML system. 
Though sUAS-based AI/ML systems have been an active area of interest, as seen with the FloodNet \cite{rahnemoonfar2021floodnet} and RescueNet \cite{rahnemoonfar2023rescuenet} competitions, this work differs via the use of the largest labeled post-disaster sUAS imagery for building damage assessment, operator training for ML-assisted damage assessment, and the operational deployment of the developed systems in response to two federally declared disasters. Furthermore, this work is motivated by the operational use of satellite-based AI/ML systems but differs in the use of sUAS imagery and operator training. 

\begin{figure}
    \centering
    \includegraphics[width=\linewidth]{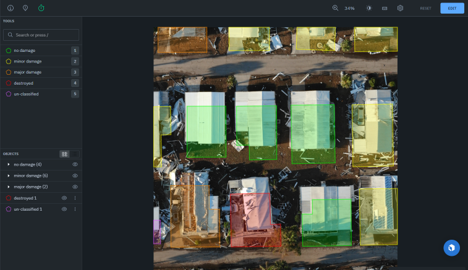}
    \caption{The LabelBox interface used by 130 annotators to label all buildings in the CRASAR-U-DROIDs dataset.}
    \label{fig:annotation_tool}
\end{figure}

\subsection{Building Damage Assessment in sUAS Imagery}
While automated sUAS-based building damage assessment has been an active area of research, prior efforts have been limited to academic settings without documented operational deployments. These prior efforts have been developed in conjunction with post-disaster sUAS imagery datasets\cite{rahnemoonfar2021floodnet, rahnemoonfar2023rescuenet, pi2020convolutional, cheng2021dorianet, zhu2021msnet}. However, these prior datasets pose limitations that would hinder operational deployment. 

There have been five datasets containing post-disaster sUAS imagery for building damage assessment \cite{rahnemoonfar2021floodnet, rahnemoonfar2023rescuenet, pi2020convolutional, cheng2021dorianet, zhu2021msnet}, but they are limited in imagery sources or disaster types and use ad-hoc labeling schemas. Only two out of the five datasets, FloodNet \cite{rahnemoonfar2021floodnet} and RescueNet \cite{rahnemoonfar2023rescuenet}, contain imagery sourced solely from operational deployments of sUAS at disasters. The remaining datasets source their imagery from ``YouTube" \cite{pi2020convolutional, cheng2021dorianet} and ``Social Media Platforms"  \cite{zhu2021msnet}, hindering the operational use of any downstream models. Taking FloodNet and RescueNet, these two datasets are limited in the scope of damage types covered and their size. For instance, FloodNet contains 28.86 gigapixels of imagery from a single disaster, Hurricane Harvey, and RescueNet contains 53.99 gigapixels of imagery from a single disaster, Hurricane Michael. These pose limitations for downstream models as they may be unable to generalize to new damage types (e.g., fire damage) and disasters (e.g., tornadoes). 

\subsection{Operational Deployments of Automated Satellite-based Building Damage Assessment}
There are two known instances of formally documented operational deployment for satellite-based building damage assessment models at disasters.  These two instances are the Mississippi Rolling Fork Tornado outbreak \cite{robinson2023rapid}, and the 2023 Turkey Earthquake \cite{robinson2023turkey}. In the Mississippi Rolling Fork Tornado outbreak ``a U-Net with an ImageNet pre-trained ResNet-50 backbone" was used \cite{robinson2023rapid}, and in the case of the 2023 Turkey Earthquake, a ``convolutional neural network" was used \cite{robinson2023turkey}. It is likely that other, undocumented instances of such model deployments exist, as similar models are reportedly in use by the United States Defense Innovation Unit \cite{xView2challenge}, United Nations Satellite Center \cite{UNOSAT_2025}, and Microsoft's AI for Good Lab \cite{robinson2023rapid, robinson2023turkey}. While other alternative architectures exist, notably with ``more than 2000 submissions" being made to the initial xBD competition, there appears to be no known public documentation indicating operational use of other model architectures \cite{xView2challenge}. While it would be convenient to deploy the ML models already developed for satellite imagery on sUAS imagery, previous work has shown that this would be insufficient. First, there would be meaningful performance degradation if imagery with a spatial resolution other than that observed during training were passed to models not adapted for high-resolution sUAS imagery \cite{reed2023scale, manzini2025challenges}. Second, if the imagery was downsampled to a spatial resolution that would be usable for such models, the underlying distribution of damage represented in the imagery would be significantly altered, lowering the utility of the collected imagery and potentially misrepresenting the conditions in the scene to decision makers \cite{manzini2025now}. While deploying such models on sUAS imagery is possible, a dedicated sUAS solution is necessary.

\section{Data Curation}

Following Fig. \ref{fig:deployment_cycle}, the first stage in the development process is data curation. This step leverages the newly released CRASAR-U-DROIDs dataset, the JDS labeling schema \cite{gupta2019creating}, and a formal two-stage review process.

The first step in data curation is the initial imagery curation, which relies on the historical deployments of the Center for Robot-Assisted Search and Rescue to ten federally declared disasters where orthomosaics were collected \cite{manzini2024crasar}. 
The resultant imagery consisted of 52 orthomosaics, and a total of 67.13 gigapixels of imagery, from six Hurricanes (Harvery, Michael, Laura, Ida, Ian, Idalia), one wildfire (Musset Bayou Fire), one tornado outbreak (Mayfield Tornado), one volcano eruption (Kilauea Volcanic Eruption), and one building collapse (Champlain Towers Collapse) \cite{manzini2024crasar}. 

The second step in data curation is building damage annotation. 21,716 buildings were labeled by 130 annotators according to the Joint Damage Scale (JDS), resulting in the largest dataset of post-disaster sUAS imagery and building damage assessment labels, the CRASAR-U-DRIODs dataset \cite{manzini2024crasar}.  
During initial imagery curation, the 52 orthomosaics were tiled into 2048x2048 images, which were uploaded to the annotation tool, Labelbox \cite{Labelbox2024}. 130 annotators were instructed to annotate the image tiles according to JDS \cite{gupta2019creating}. An example of an annotated image tile is shown in Fig. \ref{fig:annotation_tool}. The JDS was selected to align with practice, as the schema was developed with consultations from federal agencies, such as the Federal Emergency Management Agency. These annotations were reviewed via a 2-stage process detailed in \cite{manzini2024crasar}, forming the largest known post-disaster sUAS building damage assessment dataset. 

\section{Approach}
\label{sec:approach}
The approach taken in this work is a sociotechnical one intended to address the technical AI/ML problem of automated building damage assessment from sUAS imagery and the human operator component, as undoubtedly humans will need to interpret the outputs of the AI/ML system to make decisions during response and recovery operations.  
This approach corresponds to the Model Development and Operator Training stages in Fig. \ref{fig:deployment_cycle}. The problem is framed as a semantic segmentation problem for which eight models were trained and compared. UNet with Attention was the most performant for the two metrics, macro F1 aligned and unaligned, providing the community with a meaningful baseline to leverage for future work. Informal and formal operator training revealed gaps between what and how responders wanted results communicated to them and traditional AI/ML data products and expressions of accuracy, providing a better understanding of the human factors in using AI/ML.

\subsection{Model Development}
This work frames building damage assessment as an image segmentation task, where pixels associated with buildings are labeled according to damage levels, consistent with prior academic work \cite{robinson2023rapid, gupta2019xbd, cheng2021dorianet, zhu2021msnet, rahnemoonfar2023rescuenet, rahnemoonfar2021floodnet}.  
Individual buildings in sUAS imagery are described by pre-defined building polygons, and pixels under those polygons are labeled according to their damage class. To coalesce the many pixel labels to a single building label, the pre-defined building polygons are used to mask the labeled pixels, and pixels under the building polygon mask are consolidated via summation to provide a label for the entire building. During training, models are trained to label pixels, but during validation and testing, models are measured based on their ability to label buildings. This section of the document corresponds to the ``Model Development" phase depicted in Fig. \ref{fig:deployment_cycle}.

\begin{table*}[ht]
    \resizebox{\linewidth}{!}{%
    \centering
    \begin{tabular}{|ll|c|c|}
\hline
\multicolumn{2}{|c|}{\textbf{Encoder-Decoder Architectures}}                                                                       & {\textbf{Macro F1}}                     & {\textbf{Macro F1}} \\ \cline{1-2}
\multicolumn{1}{|c|}{\textbf{Encoder}}                                          & \multicolumn{1}{c|}{\textbf{Decoder}}          & \multicolumn{1}{c|}{\textbf{(Aligned)}} & \multicolumn{1}{c|}{\textbf{(Unaligned)}} \\ \hline
\multicolumn{1}{|l|}{\textbf{ResNet101} \cite{he2016deep}}                      & \textbf{DeepLabv3Plus} \cite{chen2018encoder}  & 0.493                                   & 0.475                                     \\
\multicolumn{1}{|l|}{\textbf{ViT-L ScaleMAE (Pretrained)} \cite{reed2023scale}} & \textbf{UperNet} \cite{xiao2018unified}        & 0.372                                   & 0.356                                     \\
\multicolumn{1}{|l|}{\textbf{ResNet101} \cite{he2016deep}}                      & \textbf{PSPNet} \cite{zhao2017pyramid}         & 0.377                                   & 0.368                                     \\
\multicolumn{1}{|l|}{\textbf{ViT-L ScaleMAE} \cite{reed2023scale}}              & \textbf{UperNet} \cite{xiao2018unified}        & 0.309                                   & 0.303                                     \\
\multicolumn{1}{|l|}{\textbf{ViT-L ScaleMAE (Pretrained)} \cite{reed2023scale}} & \textbf{Segmenter} \cite{strudel2021segmenter} & 0.174                                   & 0.163                                     \\
\multicolumn{1}{|l|}{\textbf{ViT-L ScaleMAE} \cite{reed2023scale}}              & \textbf{Segmenter} \cite{strudel2021segmenter} & 0.119                                   & 0.119                                     \\ \hline
\multicolumn{2}{|c|}{\textbf{Standalone Architectures}}                                                                          &  \cellcolor{gray}                                       & \cellcolor{gray}                                          \\ \hline
\multicolumn{2}{|c|}{\textbf{Attention UNet} \cite{oktay2018attention}}                                                          & \textbf{0.554}                          & \textbf{0.524}                            \\
\multicolumn{2}{|c|}{\textbf{Vanilla UNet} \cite{ronneberger2015u}}                                                              & 0.531                                   & 0.498                                     \\
\multicolumn{2}{|c|}{\textbf{Random Baseline}}                                                                                   & 0.175                                   & 0.175                                     \\ \hline
\end{tabular} }
    \caption{The performance of the 8 trained models on the CRASAR-U-DROIDs test set, grouped by architecture type and ordered by Macro F1 (Aligned). A random baseline is provided for reference. Highest values in each column are in bold.}
    \label{tab:baseline_models}
\end{table*}

\subsubsection{Model Training}
Eight baseline models were trained and evaluated using the CRASAR-U-DRIODs dataset, with the UNet with Attention being the highest performing model.  
As shown in Table \ref{tab:baseline_models}, the baseline models consisted of UNet with Attention \cite{oktay2018attention}, Vanilla UNet (UNet w/out Attention) \cite{ronneberger2015u}, ResNet101 \cite{he2016deep} + DeepLabV3Plus \cite{chen2018encoder}, ResNet101 \cite{he2016deep} + PSPNet \cite{zhao2017pyramid}, ViT-L \cite{reed2023scale} + Segmenter \cite{strudel2021segmenter}, ViT-L(Pretrained) \cite{reed2023scale} + Segmenter \cite{strudel2021segmenter}, ViT-L \cite{reed2023scale} + UperNet \cite{xiao2018unified}, and ViT-L(Pretrained) \cite{reed2023scale} + UperNet \cite{xiao2018unified}. These models represented the subject of the efforts to develop a performant model for building damage assessment in sUAS imagery.

The eight baseline models were trained on 40 orthomosaics from six disasters (Hurricanes Ian, Ida, Laura, Harvey, Kilauea Volcano Eruption, and Champlain Towers Collapse) containing 17,624 building labels. 
During training, buildings from the orthomosaics were sampled via a weighted sampler, which sampled image tiles so the distribution between the five damage classes would be approximately uniform, and training was concluded after 36 hours. During training, all models were trained using pixel-level categorical cross entropy\footnote{Reference the supplementary materials for training details.}.
All models were validated using macro F1 based on four held-out orthomosaics. Training was completed after 36 hours, and for each architecture, the model with the highest validation macro F1 was evaluated. 

\subsubsection{Model Evaluation}
Following model training, the eight baseline models were evaluated on 12 orthomosaics from four disasters (Hurricanes Michael, Idalia, Musset Bayou Fire, and Mayfield Tornado) containing 4,092 buildings, with the UNet with attention being the most performant with a macro F1 of 0.554. 
To determine the effectiveness of a model when deployed on an unseen disaster, a disaster-based train/test split, also detailed in \cite{manzini2024crasar}, was constructed to evaluate the ability of new models to generalize to ``new" disasters. 
Further, macro F1 was employed to evaluate each model's predictions across all five damage classes. As the literature notes, predefined building polygons may not align with aerial imagery \cite{manzini2024non, zampieri2018multimodal, vargas2019correcting}. To account for this, evaluations were performed on both aligned and unaligned building polygons. Details on alignment in CRASAR-U-DROIDs can be found in \cite{manzini2024non} and \cite{manzini2024crasar}. The resultant performances are shown in Table \ref{tab:baseline_models}. As the Attention UNet model was the most performant, it was selected for deployment.

\subsection{Operator Training}

Following informal training with Florida-based agencies during the initial deployment, described in the following section, 91 Disaster Practitioners from 57 Agencies were formally trained, representing broad interest and providing further feedback on the Automated Building Damage Assessment System.  
This training focused on the benefits and limitations of aerial imagery and ML-derived model outputs, the expected behavior of the AI/ML system and the observed error directionality, the historical inference times, data product formats, recommendations for sUAS flight plans and imagery capture, and finally, points of contact for requesting the deployment.  
Of the 91 practitioners who attended training, 21 completed a post-training survey and 43\% (n=9) responded ``yes" when questioned if they anticipate utilizing the AI/ML system during the 2025 Atlantic Hurricane Season, 52\% (n=11) ``maybe," and 5\% (n=1) ``no."

Surprisingly, interest by practitioners in model accuracy was low. 
Instead, practitioners focused on ensuring that the outputs of the AI/ML system could be integrated into their existing workflows and were consistent with data products already in use by the practitioner organizations. Only one question from practitioners focused on model evaluation. 

Interest remains high and, at the time of writing, a second formal training has been scheduled for September 2025.  
This training will be oriented toward the deployment of the AI/ML system during the 2025 Atlantic Hurricane Season.

\section{Deployment at Hurricanes Debby \& Helene}
\label{sec:deployment_hurricanes}

\begin{figure}
    \centering
    \includegraphics[width=\linewidth]{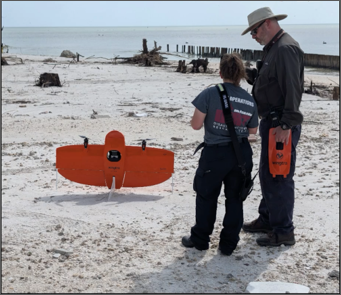}
    \caption{A Wingtra WingtraOne sUAS prepares for flight over Dekle Beach, FL, following Hurricane Helene to capture imagery. Photo credit FL-UAS1.}
    \label{fig:helene_flight}
\end{figure}

Finally, corresponding to the ``deployment" phase in Fig. \ref{fig:deployment_cycle}, is this work's deployment of the overall system at Hurricanes Debby and Helene, representing the first known operational deployment of a sUAS-based ML model for building damage assessment \cite{manzini2025challenges}.  
Both Hurricanes saw the operational deployment of the Attention UNet for automated damage assessment on sUAS imagery collected by the Florida State Response team task force FL-UAS1. All sorties were flown by FL-UAS1 using a Wintra WingtraOne sUAS, shown in Fig. \ref{fig:helene_flight}. Inference was performed by remote members of the research team, and all model outputs were manually inspected prior to delivery to emergency managers. Across both Hurricanes, 415 buildings were assessed with inference taking 18 minutes total.

The AI/ML system inferred 14.10 GB of imagery and 222 buildings in Suwannee County, Florida, following Category 1 Hurricane Debby. The buildings were located in rural residential areas. Given the storm's relatively low intensity, the buildings received lower damage classifications. 

The AI/ML system inferred 7.025 GB of imagery and 193 buildings in Lafayette and Taylor counties in Florida following Category 4 Hurricane Helene. Similar to Suwannee County, the buildings assessed were in rural residential areas. However, the storm's intensity was relatively higher, with buildings receiving higher damage classifications. An example of Hurricane Helene imagery is shown in Fig. \ref{fig:sample_imagery}.

From the deployment of the developed model at Hurricanes Debby and Helene, four challenges were observed within practice, as detailed in \cite{manzini2025challenges}. 
\begin{enumerate}
    \item The varying resolutions of the sUAS imagery collected, ranging from 1.65cm/px to 25.3cm/px, contributed to model performance degradations. 
    \item Misalignment between spatial data and sUAS imagery degraded model performance by at least 39.1\%, requiring manual adjustments, increasing data-to-decision times.
    \item Limited wireless connectivity increased the data-to-decision times due to increased latency for transmitting the collected sUAS imagery to remote research teams to deploy the model and for transmitting the model outputs and data products to teams at the disaster.
    \item Model output formats were initially insufficient for decision-makers, requiring further iterations to provide appropriate formats (e.g., GEOJSON, CSV, and compatible formats for commonly used tools such as ArcGIS), consistent with operator training discussed earlier.
\end{enumerate}

\section{Path to Future Deployment}

The path to future deployment requires engagement from both academic and operational partners to enable a continuous feedback loop between the future deployment of models, the grounded measurements of the environment in which they are deployed, and their integrations with human operators. 
This path has three elements, and they are as follows.

\begin{enumerate}
    \item Establish new performance metrics: Current measures for success used to train and evaluate ML systems are not aligned with operator concerns, and so new metrics are required to bridge that gap. 
    During deployments and operator training, disaster practitioners were less interested in model accuracy than anticipated. Instead, operators focused on advancing decision-making times and decision quality. This represents a fundamental misalignment with practice and suggests better metrics are needed.
    \item Integration of Ground Level Building Damage Assessments: Past work has shown that there are inconsistencies between imagery-derived labels and labels derived from ground-level assessments \cite{manzini2025now}. 
    For future iterations of this system to be trusted in practice, the agreement between aerial imagery-derived labels and labels derived from ground-level assessments must be quantified, and the model must be augmented to ensure that labels align with actual disaster conditions.
    \item Human Operator Training \& Feedback: In its current form, the AI/ML solution requires manual oversight from research team members. 
    This is not a sustainable solution for future deployments where research team members may not be available. While there has been success in the deployment of the models to Hurricanes Helene and Debby, the overall system needs to be hardened and packaged so that it can be deployed widely and decoupled from the manual 
    oversight of the research team.
\end{enumerate} 

\section{Future Work}
Future work should focus on improving the robustness and communication, specifically in three directions, based on the challenges discussed in the ``Initial Deployments at Hurricanes Debby \& Helene" section. 
\begin{enumerate}
    \item Autonomous Alignment of Spatial Data with Imagery: As discussed in the ``Approach" Section, deployable models must handle spatial misalignment to avoid risk of performance degradation in practice.
    \item Multi-Scale Models: As discussed in the ``Deployment at Hurricanes Debby \& Helene" Section, deployable models must handle the varying resolutions with the imagery acquired and transmitted in practice.
    \item Uncertainty Quantification \& Model Outputs Communication: As discussed in the ``Deployment at Hurricanes Debby \& Helene" Section, human oversight is currently required to monitor model behavior; however, in the future, models need to assist with this process and communicate when errors are likely to occur, especially when these systems are deployed at larger disaster events.
\end{enumerate}

\section{Ethical Considerations}
The disparate impacts of disasters are well known \cite{reid2013disasters}, and these considerations remain for AI systems \cite{gevaert2021fairness}. 
As disasters are by definition breakthrough events, they naturally generate out-of-distribution data, which risks undefined model behavior that may exacerbate disparate impacts. 
Although technical work to mitigate these impacts in AI systems is ongoing, the assumption must be that the risks are always present, and so, in the near term, human oversight becomes the only practical safeguard.
As a result, during all deployments in this work, oversight was maintained by a team with technical expertise in the overall system and practical experience in disaster response.

Any AI application for disaster management raises larger ethical issues of equity by emergency managers beyond concerns over algorithmic trustworthiness. Note that in this use case, publicly accountable emergency managers use agency-specific protocols and predictive hazard models (e.g., FEMA Flood Maps) for tasking sUAS data collection as part of their data-to-decision chain. The presented AI/ML system is one element of that chain, with its value being that it has been trained, tested, and applied to imagery acquired during actual response operations. An assessment of whether the managers’ sUAS tasking and decisions reflect societal biases or other inequalities is outside this work’s purview. 

\section{Conclusion}
This work documented the development and initial deployment of an AI/ML system for rapid damage assessment from sUAS imagery, enabling future operational deployments. 
It outlines a deployment cycle that integrates academic work with operational expertise in emergency management. This work has contributed the curation of sUAS imagery from 10 federally declared disasters, the labeling of 21,716 buildings, the training of eight baseline models for building damage assessment and the training of 91 disaster practitioners for operationalization of the overall system, and culminated in the first operational deployment of an automated sUAS damage assessment system benefiting the AI/ML, remote sensing, and emergency management communities.

\section*{Acknowledgments}
This work is supported by the AI Research Institutes Program funded by the National Science Foundation under the AI Institute for Societal Decision Making (NSF AI-SDM), Award No. 2229881, and under ``Datasets for Uncrewed Aerial System (UAS) and Remote Responder Performance from Hurricane Ian" Award No. 2306453. 
The authors thank the Florida State Emergency Response Team, FL-UAS1 task force, and Florida State University for their support.

\bibliography{aaai2026}

\end{document}